\documentclass[review]{elsarticle}

\usepackage{hyperref}

\journal{Journal of \LaTeX\ Templates}

\usepackage{hyperref}
\usepackage[utf8]{inputenc} 
\usepackage{todonotes,multirow}
\usepackage{amsmath}
\usepackage{algorithm}
\usepackage{algorithmic}
\usepackage{amssymb}
\usepackage{booktabs}
\usepackage{color}
\usepackage{makecell}
\usepackage{multirow}
\usepackage[caption=false]{subfig}
\usepackage{textcomp} 
\usepackage{gensymb} 
\newcommand{\eg}{e.g.}
\newcommand{\ie}{i.e.}
\newcommand{\etal}{\textit{et al.~}}

\begin{document}

\begin{frontmatter}

\title{Low-rank representations with incoherent dictionary for face recognition}
\tnotetext[mytitlenote]{Fully documented templates are available in the elsarticle package on \href{http://www.ctan.org/tex-archive/macros/latex/contrib/elsarticle}{CTAN}.}

\author[add1]{Pei Xie}
\author[add2,add3]{He-Feng Yin}
\author[add2,add3]{Xiao-Jun Wu\corref{mycorrespondingauthor}}
\cortext[mycorrespondingauthor]{Corresponding author}
\ead{wu\_xiaojun@jiangnan.edu.cn}

\address[add1]{Jiangsu GuoGuang Electronic Information Technology Co., Ltd., Changzhou, 213015, China }
\address[add2]{School of Internet of Things Engineering, Jiangnan University, Wuxi 214122, China}
\address[add3]{Jiangsu Provincial Engineering Laboratory of Pattern Recognition and Computational Intelligence, Jiangnan University, Wuxi 214122, China}

\begin{abstract}
Face recognition remains a hot topic in computer vision, and it is challenging to tackle the problem that both the training and testing images are corrupted. In this paper, we propose a novel semi-supervised method based on the theory of the low-rank matrix recovery for face recognition, which can simultaneously learn discriminative low-rank and sparse representations for both training and testing images. To this end, a correlation penalty term is introduced into the formulation of our proposed method to learn an incoherent dictionary. Experimental results on several face image databases demonstrate the effectiveness of our method, i.e., the proposed method is robust to the illumination, expression and pose variations, as well as images with noises such as block occlusion or uniform noises.
\end{abstract}


\end{frontmatter}


\section{Introduction}
Face recognition (FR) is an interesting and popular issue in the communities of computer vision and pattern recognition, which plays an important role in practical applications, \eg, security monitoring and access control. Reducing the dimensionality of the face images is a simple yet effective idea for FR, classical methods like Eigenfaces~\cite{turk1991face}, Fisherfaces~\cite{belhumeur1997eigenfaces}, Laplacianfaces~\cite{he2005face} and their extensions~\cite{xiao2004new,zheng2006nearest,zheng2006reformative}. With the efforts of the researchers, local discriminant embedding (LDE)~\cite{chen2005local} and many other methods have been successfully applied in FR. On controlled datasets many face recognition algorithms can achieve superb performance. Unfortunately, they cannot perform well in some real-world situations. According to~\cite{wagner2011toward} and~\cite{de2001robust}, the key point to solve the problem is to handle variations in illumination, images misalignment, and occlusion in the testing images simultaneously.

In recent years, the sparse-representation-based classification (SRC)~\cite{wright2008robust} has become very popular in FR. The key idea of SRC is to represent an input sample as a sparse linear combination over all the training samples, in which the $\ell_1$-norm is usually used as a sparsity constraint. The results in~\cite{wright2008robust} on FR are promising, and SRC is also robust to face occlusion and corruption in the test images. Nevertheless, the training images should be well aligned in SRC. To deal with face misalignment, an improved method is presented in~\cite{wagner2011toward}. To counteract outliers such as noises in FR, Yang proposed a modified SRC-based framework~\cite{yang2010gabor,yang2011robust}. Unfortunately, the above methods cannot handle well the situation when both training and testing samples are corrupted. To further assess the robustness of sparse representation based recognition algorithms, this problem should be addressed. 

According to traditional subspace theory, face images of the same class are often lie in the same low-rank subspace~\cite{liu2012robust,basri2003lambertian,hastie1998metrics}, but due to illumination changes, expression or pose variations and other issues, the actual face images often do not maintain the low-rank structure, which will affect the recognition performance. Recently, low-rank matrix recovery (LRMR), has been successfully applied to FR~\cite{liu2012robust}, LRMR can obtain the low-rank data even from the corrupted data. It can efficiently remove sparse noises like illumination changes and occlusions in corrupted face images. Yin \etal\cite{yin2016face} presented a new method called low rank matrix recovery with structural incoherence and low rank projection (LRSI\_LRP) which can correct the corrupted test images with a low rank projection matrix. Zhao \etal\cite{zhao2015collaborative} developed a discriminative low-rank representation method for collaborative representation-based (DLRR-CR) robust face recognition. Chen \etal\cite{chen2018robust} proposed a robust low-rank recovery algorithm (RLRR) with a distance-measure structure for face recognition. For the recognition problem, label information should be incorporated to learn discriminative dictionary. Chen \etal\cite{chen2012low} adopts the low-rank recovery technique to recover the clean low-rank data matrix from the training data class by class and a structural incoherence term is introduced to enforce the resulting low-rank dictionary for each class to be independent, then the clean data matrix is employed as dictionary for SRC to classify test data. Ma \etal\cite{ma2012sparse} proposed a discriminative low-rank dictionary learning method for FR. The proposed algorithm tries to optimize the sub-dictionaries for each class to be low-rank. In order to learn a dictionary which is not only low-rank but also discriminative, Li \etal\cite{li2013discriminative} applied the constraint on the coding coefficients to make the ratio of the within-class scatter to the between-class scatter be small. However, all of them explore structural information class by class, which cannot capture the global structure. To explore the global structure information among all the training images, Zhang \etal\cite{zhang2013learning} proposed a discriminative, structured low-rank method. It incorporates a code regularization term with block-diagonal structure which can regularize images from the same class to have the same representation to learn discriminative dictionary. However, it is not usually the case. The fact is that face images which are from the same class usually lie in the same subspace and have relevant representation, but the representation is not necessarily the same. To overcome the defect, Li \etal\cite{li2014learninga,li2014learningb} proposed a semi-supervised framework to learn robust face representations with classwise block-diagonal structure. It can learn robust representations of training and testing images simultaneously and capture classwise block-diagonal structure with a classwise regularization term at the same time. It can achieve significant performance gains even when both the training and testing samples are corrupted.

Prior work demonstrated that high-quality dictionary can greatly improve the performance of sparse representation approaches~\cite{yang2010gabor,yang2011robust,liu2012robust,chen2012low}. The re-searchers have made many efforts to learn a good dictionary~\cite{chen2012low,ma2012sparse,li2013discriminative}. The dictionary learning techniques always focus on the reconstruction accuracy and discriminative power of the dictionary. For example, in \cite{ma2012sparse} and \cite{li2013discriminative} the relationship between two dictionaries for different classes was addressed. Lin \etal\cite{lin2012incoherent} proposed an incoherent dictionary learning model that maximizes the incoherence of basis atoms in one single output dictionary, and the experimental results demonstrated the advantages of incoherence constraint.

\begin{figure}[h]
\centering
\includegraphics[trim={0mm 0mm 0mm 0mm},clip, width = .8\textwidth]{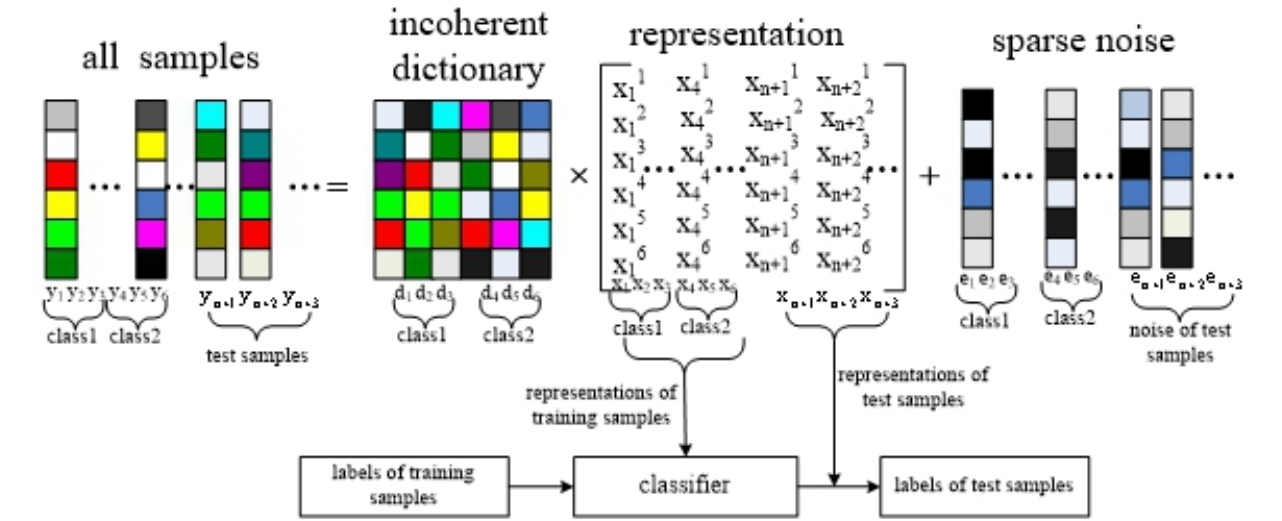}
\caption{Schematic diagram of our proposed method. Representations of the training and test images are simultaneously learned, and a simple yet effective linear classifier is learned based on the representation matrix and label matrix of the training data. By imposing the incoherence constraint on the dictionary, an incoherent dictionary can be learned in the training process of our proposed method..}
\label{fig:flowchart}
\end{figure}

\begin{figure}[h]
\centering
\includegraphics[trim={0mm 0mm 0mm 0mm},clip, width = .8\textwidth]{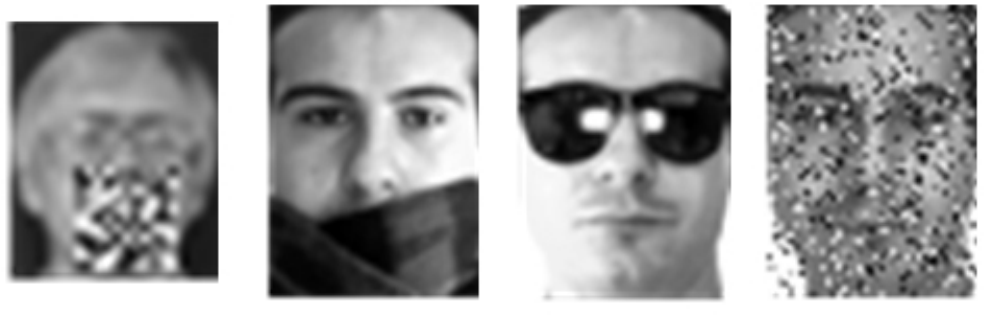}
\caption{Corrupted face images, the first one is occluded by an unrelated random images, the middle two images are real disguises, and the last image is contaminated by random pixel corruption.}
\label{fig:img_corr}
\end{figure}

In this paper, a novel semi-supervised method based on the theory of low-rank matrix recovery is proposed for FR, which focuses on handling the situation that both the training and testing samples are corrupted. Meanwhile, inspired by \cite{lin2012incoherent} we explicitly incorporate a correlation penalty in the dictionary learning model to make the dictionary more discerning. The representation learning process is shown in Fig.~\ref{fig:flowchart}. Our method has several advantages which are summarized as follows.
\begin{itemize}
\item We adopt a semi-supervised framework to learn face representation with incoherent dictionary, which can simultaneously learn low-rank and sparse representations of both training and testing images. 
\item Furthermore, our proposed method is robust to the case that both training and testing images are badly corrupted. We conducted some experiments under the scenarios where samples are corrupted like in Fig.~\ref{fig:img_corr}.
\item In addition, incoherent dictionary with desirable reconstruction and discrimination capability can be learned by the proposed approach.
\end{itemize}
The rest of this paper is outlined as follows. First, we briefly review related work 0n the low-rank matrix recovery and incoherent dictionary learning in Section~\ref{sec_2}. Section~\ref{sec_3} presents our method. The optimization process based on the inexact Augmented Lagrange Multiplier method for our model is given in Section~\ref{sec_4}. Section~\ref{sec_5} illustrates the classification technique. Computational complexity of our proposed method is presented in Section~\ref{sec_6}. Experimental results on several face image databases and corresponding discussions are given in Section~\ref{sec_7}. Finally, Section~\ref{sec_8} concludes this paper.

\section{Related Work}
\label{sec_2}
First, we summarize some notations. Throughout this paper, notations with a bar denote symbols of training images, while notations with a hat denote symbols of testing images. Let $\bar{\mathbf{Y}}=\left [ \bar{\mathbf{Y}}_{1},\bar{\mathbf{Y}}_{2},\ldots ,\bar{\mathbf{Y}}_{C} \right ]\in \mathbb{R}^{d\times n}$ be the feature matrix which is composed of training images of $C$  classes, and the feature matrix $\hat{\mathbf{Y}}=\left [ \hat{\boldsymbol{y}}_{1},\hat{\boldsymbol{y}}_{2},\ldots,\hat{\boldsymbol{y}}_{n_s} \right ]\in \mathbb{R}^{d\times n_{s}}$ is composed of $n_s$ testing images. Thus $\mathbf{Y}=\left [\bar{\mathbf{Y}}, \hat{\mathbf{Y}} \right ]$ is the feature matrix of the whole database, where each column of $\mathbf{Y}$ denotes the feature of an image. Each image of $\mathbf{Y}$ can be coded as a linear combination of the base elements of dictionary $\mathbf{D}\in \mathbb{R}^{d \times m}$, \ie
\begin{equation}
\label{eq:linear_rep}
\mathbf{Y}=\mathbf{D}\mathbf{X}
\end{equation}
where $\mathbf{X}=\left [\bar{\mathbf{X}}, \hat{\mathbf{X}} \right ]$, and $\bar{\mathbf{X}}$ corresponds to the representation matrix of training images, while $\hat{\mathbf{X}}$ corresponds to the representation matrix of testing images.  

\subsection{Low-Rank and Sparse Representation}
Suppose we are given a set of corrupted images from multiple classes, now the problem is to learn robust representations of training and testing images simultaneously.
Low-rankness is an appropriate criterion to capture the low dimensional structure from the high-dimensional data \cite{liu2012robust}. To obtain the clean database $\mathbf{Y}$, learning low-rank representations can be solved as follows, 
\begin{equation}
\label{eq:lrr_formu}
\underset{\mathbf{X},\mathbf{E}}{\textrm{min}} \ \textrm{rank}\left ( \mathbf{X} \right )+\lambda \left \| \mathbf{E} \right \|_0, \ \textrm{s.t.} \  \mathbf{Y}=\mathbf{D}\mathbf{X}+\mathbf{E}
\end{equation}
                                                 
Effectiveness of low-rank and sparse representation for the classification problem has been demonstrated in \cite{chen2012low} and \cite{mairal2010online}.

As we know from \cite{liu2012robust}
 and \cite{wright2009robust}, low-rank model can reveal the subspace structure of data, sparse representation has been shown to achieve promising performance in FR \cite{wright2008robust,yang2009linear,tang2009inferring}. Taking advantage of low-rank and sparse representation is very meaningful, and low-rank and sparse representation for the classification problem is very effective \cite{zhang2013learning,zhang2013low}. Li \etal\cite{li2014learningb} imposed both low-rank and sparse constraints on the representation matrix, the method is called learning low-rank and sparse representation (LRRS) for FR, this model is formulated as follows,
\begin{equation}
\label{eq:lrsr_ori}
\underset{\mathbf{X},\mathbf{E}}{\textrm{min}} \ \textrm{rank}\left ( \mathbf{X} \right )+\lambda\left \| \mathbf{E} \right \|_{0}+\beta\left \| \mathbf{X} \right \|_{0}  \  \textrm{s.t.} \ \mathbf{Y}=\mathbf{D}\mathbf{X}+\mathbf{E}
\end{equation}                                
As the above problem is non-convex, the common way is to substitute the $\ell_1$-norm for $\ell_0$-norm and the nuclear norm for the rank function. Therefore, the optimization problem (\ref{eq:lrsr_ori}) is relaxed into the following problem,
\begin{equation}
\label{eq:lrsr_formu}
\underset{\mathbf{X},\mathbf{E}}{\textrm{min}} \ \left \| \mathbf{X} \right \|_*+\lambda\left \| \mathbf{E} \right \|_{1}+\beta\left \| \mathbf{X} \right \|_{1}  \  \textrm{s.t.} \ \mathbf{Y}=\mathbf{D}\mathbf{X}+\mathbf{E}
\end{equation}                                  
The nuclear norm $\left \| \mathbf{X} \right \|_{*} $  (\ie, the sum of singular values), $\left \| \cdot \right \|_{1} $ is defined as the sum of the absolute values of entries in the given matrix, $\lambda$ and $\beta$ control the sparsity of sparse noise term  $\mathbf{E}$ and sparse representation term $\mathbf{X}$, and $\mathbf{D}$ is the dictionary that linearly spans the data space. 

\subsection{Incoherent Dictionary Learning (IDL) model}
The quality of dictionary is critical for FR, especially when training and testing samples are corrupted, high-quality dictionary can greatly improve the performance of sparse representation approaches\cite{yang2010gabor,yang2011robust,liu2012robust,chen2012low}. Lin \etal\cite{lin2012incoherent} proposed an incoherent dictionary learning model, which maximizes the incoherence of basis atoms in one single output dictionary to learn a discriminative dictionary. The correlation measure of a dictionary $\mathbf{D}$ can be defined as,
\begin{equation}
\label{eq:inco_term}
\textrm{cor}\left ( \mathbf{D} \right )=\left \| \mathbf{D}^{T}\mathbf{D}-\mathbf{I}\right \|_{F}^{2}
\end{equation}
where $\mathbf{I}\in R^{m \times m} $  is an identity matrix, we can see from the formula~(\ref{eq:inco_term}) that the correlation measure is zero when the columns of dictionary $\mathbf{D}$ are orthogonal, in which case the dictionary $\mathbf{D}$ is the most irrelevant.

\section{Low-Rank Representations with Incoherent Dictionary}
\label{sec_3}
The experimental results in \cite{li2014learninga} and \cite{li2014learningb} verified that the semi-supervised learning method can effectively deal with the case that both the training and testing samples are corrupted. Inspired by the work of \cite{li2014learninga} and \cite{li2014learningb}, we also exploit the semi-supervised learning framework. We propose a new method of low-rank representations with incoherent dictionary (LRRID) for FR. Finally, the objective function is proposed to learn the sparse and low-rank representation with the incoherent dictionary by explicitly incorporating a correlation penalty into the dictionary learning model, 
\begin{equation}
\label{eq:obj_pro}
\underset{\mathbf{D},\mathbf{X},\mathbf{E}}{\textrm{min}} \ \left \| \mathbf{X} \right \|_*+\lambda\left \| \mathbf{E} \right \|_{1}+\beta\left \| \mathbf{X} \right \|_{1}  +\gamma\left \| \mathbf{D}^{T}\mathbf{D}-\mathbf{I}\right \|_{F}^{2}, \ \  \textrm{s.t.} \ \mathbf{Y}=\mathbf{D}\mathbf{X}+\mathbf{E}
\end{equation}                                 
There are three regularization parameters ($\lambda>0$, $\beta>0$ and $\gamma>0$) that reflect relative contributions of each term.

\section{Optimization}
\label{sec_4}
By introducing two auxiliary variables $\mathbf{J}$ and $\mathbf{L}$, we convert problem (\ref{eq:obj_pro}) into the following equivalent optimization problem,
\begin{equation}
\label{eq:equ_form}
\begin{split}
&\underset{\mathbf{D},\mathbf{X},\mathbf{E},\mathbf{J},\mathbf{L}}{\textrm{min}} \ \left \| \mathbf{J} \right \|_{*}+\lambda\left \| \mathbf{E} \right \|_{1}+\beta\left \| \mathbf{L} \right \|_{1}+\gamma\left \| \mathbf{D}^{T}\mathbf{D}-\mathbf{I}\right \|_{F}^{2}, \\  &\textrm{s.t.} \ \mathbf{Y}=\mathbf{D}\mathbf{X}+\mathbf{E},\mathbf{X}=\mathbf{J},\mathbf{X}=\mathbf{L}   
\end{split}  
\end{equation}   
The optimization problem~(\ref{eq:equ_form}) can be solved based on the Augmented Lagrange Multiplier (ALM) method \cite{lin2010augmented}, so we adopt the inexact ALM method for efficiency in this paper. The augmented Lagrangian function of problem~(\ref{eq:equ_form}) is as follows,
\begin{equation}
\label{eq:augment}
\begin{split}
&\mathcal{L}\left ( \mathbf{J},\mathbf{X},\mathbf{L},\mathbf{E},\mathbf{D},\mathbf{T}_1,\mathbf{T}_2,\mathbf{T}_3,\mu \right )=\left \| \mathbf{J} \right \|_{*}+\lambda \left \| \mathbf{E} \right \|_{1}+\beta \left \| \mathbf{L} \right \|_{1}\\&+\gamma \left \| \mathbf{D}^{T}\mathbf{D}-\mathbf{I} \right \|_{F}^{2}+\left \langle \mathbf{T}_{1} ,\mathbf{Y}-\mathbf{D}\mathbf{X}-\mathbf{E}\right \rangle+\left \langle \mathbf{T}_{2},\mathbf{X}-\mathbf{J} \right \rangle+\left \langle \mathbf{T}_{3},\mathbf{X}-\mathbf{L} \right \rangle\\&+\frac{\mu}{2}\left (\left \| \mathbf{Y}-\mathbf{D}\mathbf{X}-\mathbf{E} \right \|_{F}^{2} +\left \| \mathbf{X}-\mathbf{J} \right \|_{F}^{2} +\left \| \mathbf{X}-\mathbf{L} \right \|_{F}^{2}\right )
\end{split}
\end{equation}
where $\mathbf{T}_1$, $\mathbf{T}_2$, and $\mathbf{T}_3$ are Lagrangian multipliers and $\mu>0$ is the penalty parameter, and $\left \langle \mathbf{A},\mathbf{B} \right \rangle= \textrm{trace}\left ( \mathbf{A}^{T}\mathbf{B} \right )$.  We solve problem (\ref{eq:augment}) iteratively by updating $\mathbf{J}$, $\mathbf{X}$, $\mathbf{L}$, $\mathbf{E}$ and $\mathbf{D}$ once at a time, and the detailed procedures are as follows.

Updating $\mathbf{J}$: Fix the other variables and solve the following problem,
\begin{equation}
\label{eq:update_J}
\begin{split}
\mathbf{J}^{k+1}&=\textrm{arg} \ \underset{\mathbf{J}}{\textrm{min}} \ \left \| \mathbf{J} \right \|_{*}+\left \langle \mathbf{T}_{2}^{k},\mathbf{X}^{k}-\mathbf{J} \right \rangle+\frac{\mu}{2}^{k}\left \| \mathbf{X}^{k}-\mathbf{J} \right \|_{F}^{2}\\ &=\textrm{arg} \ \underset{\mathbf{J}}{\textrm{min}}\, \frac{1}{\mu^{k}}\left \| \mathbf{J} \right \|_{*}+\frac{1}{2}\left \| \mathbf{J}-\left ( \mathbf{X}^{k}+\frac{\mathbf{T}_{2}^{k}}{\mu^{k}} \right ) \right \|_{F}^{2}\\ &=\mathbf{U}S_{\frac{1}{\mu^{k}}} \left [ \mathbf{\Sigma}  \right ]\mathbf{V}^{T} 
\end{split}
\end{equation}
where $\left ( \mathbf{U},\mathbf{\Sigma},\mathbf{V}^{T} \right )=\textrm{SVD}\left (\mathbf{X}^{k}+\frac{\mathbf{T}_{2}^{k}}{\mu^{k}}   \right )$ and $S_{\epsilon }\left [ \cdot  \right ]$ is the soft-thresholding (shrinkage) operator defined as follows,
\begin{equation}
\label{eq:soft_thres}
S_{\epsilon }\left [ x \right ]=\textrm{sign}\left ( x \right )\left ( \left | x \right |-\epsilon  \right )
\end{equation}
Updating $\mathbf{X}$: Fix the other variables and solve the following problem,
\begin{equation}
\label{eq:update_X}
\begin{split}
&\mathbf{X}^{k+1}=\textrm{arg} \ \underset{\mathbf{X}}{\textrm{min}}\,\,\,  \left \langle \mathbf{T}_{1}^{k} ,\mathbf{Y}-\mathbf{D}^{k}\mathbf{X}-\mathbf{E}^{k}\right \rangle+\left \langle \mathbf{T}_{2}^{k},\mathbf{X}-\mathbf{J}^{k+1} \right \rangle+\left \langle \mathbf{T}_{3}^{k},\mathbf{X}-\mathbf{L}^{k} \right \rangle\\&+\frac{\mu}{2}\left (\left \| \mathbf{Y}-\mathbf{D}^{k}\mathbf{X}-\mathbf{E}^{k}\right \|_{F}^{2} +\left \| \mathbf{X}-\mathbf{J}^{k+1} \right \|_{F}^{2} +\left \| \mathbf{X}-\mathbf{L}^{k} \right \|_{F}^{2}\right )
\end{split}
\end{equation}
The optimal solution to (\ref{eq:update_X}) is given by
\begin{equation}
\label{eq:solu_X}
\mathbf{X}^{k+1}=[(\mathbf{D}^{k})^{T}\mathbf{D}^{k}+2\mathbf{I}]^{-1}\left (  (\mathbf{D}^{k})^{T}\left ( \mathbf{Y}-\mathbf{E}^{k} \right )+\mathbf{J}^{k+1}+\mathbf{L}^{k}+\frac{(\mathbf{D}^{k})^{T}\mathbf{T}_{1}^{k}-\mathbf{T}_{2}^{k}-\mathbf{T}_{3}^{k}}{\mu}\right)
\end{equation}
Updating $\mathbf{L}$: Fix the other variables and solve the following problem,  
\begin{equation}
\label{eq:update_L}
\begin{split}
\mathbf{L}^{k+1}&=\textrm{arg} \ \underset{\mathbf{L}}{\textrm{min}} \ \beta \left \| \mathbf{L} \right \|_{1}+\left \langle \mathbf{T}_{3}^{k},\mathbf{X}^{k+1}-\mathbf{L} \right \rangle+\frac{\mu^{k}}{2}\left \| \mathbf{X}^{k+1}-\mathbf{L} \right \|_{F}^{2}\\&=\textrm{arg} \ \underset{\mathbf{L}}{\textrm{min}}\, \frac{\beta}{\mu}\left \| \mathbf{L} \right \|_{1}+\frac{1}{2}\left \| \mathbf{L}-\left ( \mathbf{X}^{k+1}+\frac{\mathbf{T}_{3}^{k}}{\mu^{k}} \right ) \right \|_{F}^{2}\\&=S_{\frac{\beta}{\mu^{k}}}\left [ \mathbf{X}^{k+1}+\frac{\mathbf{T}_{3}^{k}}{\mu^{k}} \right ]
\end{split}
\end{equation}
Updating $\mathbf{E}$: Fix the other variables and solve the following problem,  
\begin{equation}
\label{eq:update_E}
\begin{split}
\mathbf{E}^{k+1}&=\textrm{arg} \ \underset{E}{\textrm{min}}\,\,\, \lambda\left \| \mathbf{E} \right \|_{1}+\left \langle \mathbf{T}_{1}^{k} ,\mathbf{Y}-\mathbf{D}^{k}\mathbf{X}^{k+1}-\mathbf{E}\right \rangle+\frac{\mu^{k}}{2}\left \| \mathbf{Y}-\mathbf{D}^{k}\mathbf{X}^{k+1}-\mathbf{E} \right \|_{F}^{2}\\&=\textrm{arg} \ \underset{\mathbf{E}}{\textrm{min}}\, \frac{\lambda}{\mu^{k}}\left \| \mathbf{E} \right \|_{1}+\frac{1}{2}\left \| \mathbf{E}-\left ( \mathbf{Y}-\mathbf{D}^{k}\mathbf{X}^{k+1}+\frac{\mathbf{T}_{1}^{k}}{\mu^{k}} \right ) \right \|_{F}^{2}\\&=S_{\frac{\lambda}{\mu^{k}}}\left [ \mathbf{Y}-\mathbf{D}^{k}\mathbf{X}^{k+1}+\frac{\mathbf{T}_{1}^{k}}{\mu^{k}} \right ]
\end{split}
\end{equation}
Updating $\mathbf{D}$: Fix the other variables and solve the following problem, 
\begin{equation}
\label{eq:update_D}
\begin{split}
&\mathbf{D}=\textrm{arg} \ \underset{\mathbf{D}}{\textrm{min}} \ \gamma \left \| \mathbf{D}^{T}\mathbf{D}-\mathbf{I} \right \|_{F}^{2}+\left \langle \mathbf{T}_{1}^{k} ,\mathbf{Y}-\mathbf{D}\mathbf{X}^{k+1}-\mathbf{E}^{k+1}\right \rangle\\&+\frac{\mu^{k}}{2}\left \| \mathbf{Y}-\mathbf{D}\mathbf{X}^{k+1}-\mathbf{E}^{k+1} \right \|_{F}^{2}
\end{split}
\end{equation}
The optimization problem (\ref{eq:update_D}) can be solved via the first-order gradient descent method \cite{mairal2010online,hung2014efficient}:
\begin{equation}
\label{eq:Dq}
\mathbf{D}_{q+1}=\Pi_{\mathbf{D}}\left \{  \mathbf{D}_{q}-\eta \bigtriangledown F\left ( \mathbf{D} _{q}\right )\right \}
\end{equation}
where   
\begin{equation}
\label{eq:FD}
\begin{split}
&F\left ( \mathbf{D} \right )=\gamma \left \| \mathbf{D}^{T}\mathbf{D}-\mathbf{I} \right \|_{F}^{2}+\left \langle \mathbf{T}_{1}^{k} ,\mathbf{Y}-\mathbf{D}\mathbf{X}^{k+1}-\mathbf{E}^{k+1}\right \rangle\\&+\frac{\mu^{k}}{2}\left \| \mathbf{Y}-\mathbf{D}\mathbf{X}^{k+1}-\mathbf{E}^{k+1} \right \|_{F}^{2} 
\end{split}
\end{equation}
and 
\begin{equation}
\label{eq:roundF}
\begin{split}
&\bigtriangledown F\left ( \mathbf{D} _{q}\right )=4\gamma \left ( \mathbf{D}_{q}\mathbf{D}_{q}^{T}\mathbf{D}_{q}-\mathbf{D}_{q} \right )-\mathbf{T}_{1}^{k}(\mathbf{X}^{k+1})^{T}\\&+\mu^{k}\left ( \mathbf{D}_{q} \mathbf{X}^{k+1}(\mathbf{X}^{k+1})^{T}+\mathbf{E}^{k+1}(\mathbf{X}^{k+1})^{T}-\mathbf{Y}(\mathbf{X}^{k+1})^{T} \right )
\end{split}
\end{equation}
$\bigtriangledown F\left ( \mathbf{D}_{q }\right )$ is the gradient of $F\left ( \mathbf{D}_{q} \right )$  with respect to $\mathbf{D}_{q}$, parameter $\eta$ is the step size, and $\Pi_{\mathbf{D}} $ is the projection function which maps each column to the $\ell_2$-norm unit ball. More specifically, the dictionary is learned iteratively by finding the local minimum along the gradient direction based on the small step size and normalizing each column vector until convergence. The optimization process of (\ref{eq:augment}) is summarized in Algorithm \ref{alg1}.
\begin{algorithm}[t]
\begin{algorithmic}[1]
\vspace{0.03in}
\STATE \textbf{Input:} Feature Matrix $\mathbf{Y}$, parameter $\lambda$, $\beta$ and $\gamma$.

\STATE Initialize: $\mathbf{X}^0=0$, $\mathbf{J}^0=0$, $\mathbf{L}^0=0$, $\mathbf{T}_1^0=0$, $\mathbf{T}_2^0=0$, $\mathbf{T}_3^0=0$, $\mu^0=10^{-5}$, $\mu_{max}=10^8$, $\rho=1.1$, $\varepsilon=10^{-6}$, and dictionary $\mathbf{D}^0$ is initialized as the randomly selected training samples.
\WHILE {not converged}
\STATE Update $\mathbf{J}$ by (\ref{eq:update_J}).
\STATE Update $\mathbf{X}$ by (\ref{eq:solu_X}).
\STATE Update $\mathbf{L}$ by (\ref{eq:update_L}).
\STATE Update $\mathbf{E}$ by (\ref{eq:update_E}).
\STATE Update $\mathbf{D}$ by (\ref{eq:Dq}).
\STATE Update the multipliers:

$\mathbf{T}_1^{k+1}=\mathbf{T}_1^{k+1}+\mu^k(\mathbf{Y}-\mathbf{D}^{k+1}\mathbf{X}^{k+1}-\mathbf{E}^{k+1})$

$\mathbf{T}_2^{k+1}=\mathbf{T}_2^{k+1}+\mu^k(\mathbf{X}^{k+1}-\mathbf{J}^{k+1})$

$\mathbf{T}_3^{k+1}=\mathbf{T}_3^{k+1}+\mu^k(\mathbf{X}^{k+1}-\mathbf{L}^{k+1})$

\STATE Update $\mu$

$ \mu^{k+1}=\textrm{min}(\rho\mu^k,\mu_{max})$

\STATE Check the convergence conditions:

$\left \| \mathbf{Y}-\mathbf{D}^{k+1}\mathbf{X}^{k+1}-\mathbf{E}^{k+1} \right \|_{\infty}<\varepsilon$, $\left \| \mathbf{X}^{k+1}-\mathbf{J}^{k+1} \right \|_{\infty}<\varepsilon$ and $\left \| \mathbf{X}^{k+1}-\mathbf{L}^{k+1} \right \|_{\infty}<\varepsilon$
\ENDWHILE
\STATE \textbf{Output:} $\mathbf{X}$, $\mathbf{D}$, and $\mathbf{E}$.
\vspace{0.03in}
\end{algorithmic}
\caption{Solving Problem (\ref{eq:augment}) by Inexact ALM}
\label{alg1}
\end{algorithm}

\section{Classification}
\label{sec_5}
To classify the test data, we employ a linear classifier in the testing phase as in \cite{li2014learningb}. The coefficients $\bar{\mathbf{X}}$ for training samples and $\hat{\mathbf{X}}$ for testing samples can be derived through Algorithm \ref{alg1}. We can learn a linear classifier $\mathbf{W}$ based on the coefficients of training samples and their labels like \cite{zhang2012online},
\begin{equation}
\label{eq:obj_classifier}
\mathbf{W}^{*}=\textrm{arg} \ \underset{\mathbf{W}}{\textrm{min}}\left \| \mathbf{H}-\mathbf{W}\bar{\mathbf{X}} \right \|_{F}^{2}+\eta \left \| \mathbf{W} \right \|_{F}^{2}
\end{equation}
where $\mathbf{H}$ is the label matrix of training samples, $\eta$ is the weight of regularization term. Clearly, the formulation (\ref{eq:obj_classifier}) is a convex problem, which is solved directly by setting the partial derivatives with respect to $\mathbf{W}$ to zero, and the solution is given by
\begin{equation}
\label{eq:solu_classifier}
\mathbf{W}^{*}=\mathbf{H}\bar{\mathbf{X}}^{T}\left (  \bar{\mathbf{X}}\bar{\mathbf{X}}^{T}+\eta \mathbf{I}\right )^{-1}
\end{equation}
Then label for test sample $j$ is given by:
\begin{equation}
\label{eq:rule_classify}
i^{*}=\textrm{arg} \ \underset{i}{\textrm{max}} \ \mathbf{W}^{*}\hat{\mathbf{X}}_{j} 
\end{equation}
where $i^*$ is corresponding to the classifier with the largest output.

\section{Computational Complexity}
\label{sec_6}
The overall computational complexity of our method is $O\left ( md\left ( n_{t}+n_{s} \right ) +m^{3}\right ) $, where $m$ is the number of the bases, $d$ is the feature dimension, $n_t$ represents the number of training samples, and $n_s$ represents the number of testing samples. The computational complexity of our method is linear with the sample number. In comparison, the sparse representation based recognition methods in \cite{wright2008robust} and \cite{ma2012sparse} have computational complexity of $O\left ( m^{2}dn_{s}\right ) $ \cite{zhang2013low}, which are slower than our method when $n_{t}<\left ( m-1 \right )n_{s} $ and $m<dn_{s} $. Generally, the overall computational complexity of \cite{li2014learningb} is similar to that of ours.

\section{Experiments and Analysis}
\label{sec_7}
We evaluate our approach on three public databases: the ORL database \cite{samaria1994parameterisation}, the Extended Yale B database \cite{georghiades2001few,lee2005acquiring}, and the AR database \cite{martinez2001pca}. Some sample images from these databases are shown in Fig.~\ref{fig:exam_img}. We test the robustness of LRRID when both training and testing images are corrupted, including block occlusion, illumination change, pose variation, expression change, pixel corruptions and uniform noises, respectively. We compare our method LRRID with some state-of-the-art methods including its special case (LRRS), which are enumerated as follows,

(1) LRRID: It is to learn discriminative low-rank and sparse representations for both training and testing images simultaneously in a semi-supervised framework with the incoherent dictionary. The objective function is shown in formulation (\ref{eq:obj_pro}).

(2) LRRS \cite{li2014learningb}: It is to learn low-rank and sparse representations. The dictionary is composed by the training samples. The objective function is shown in formulation (\ref{eq:lrsr_formu}).

(3) RCBD \cite{li2014learningb}: It is to learn representation of training images, testing images and discriminative dictionary simultaneously in a semi-supervised framework with classwise block-diagonal structure regularization to the training images.
 
(4) DLRD\_SR \cite{ma2012sparse}: It is to learn discriminative low-rank dictionary class by class, and sparse representation based classification method is employed with respect to the learned dictionary.

(5) SRC \cite{wright2008robust}: It is to learn sparse representation with the dictionary which is composed of the training samples. Specially, SRC indicates the case that the whole training samples are used as the dictionary. SRC* indicates the case that the dictionary size is the same as ours.

It is worth noting that our method and the work of \cite{li2014learningb} both adopt the semi-supervised framework, so the experimental results of RCBD have reference value.  
\subsection{ORL Face Database }
\begin{figure}[h]
\centering
\includegraphics[trim={0mm 0mm 0mm 0mm},clip, width = .8\textwidth]{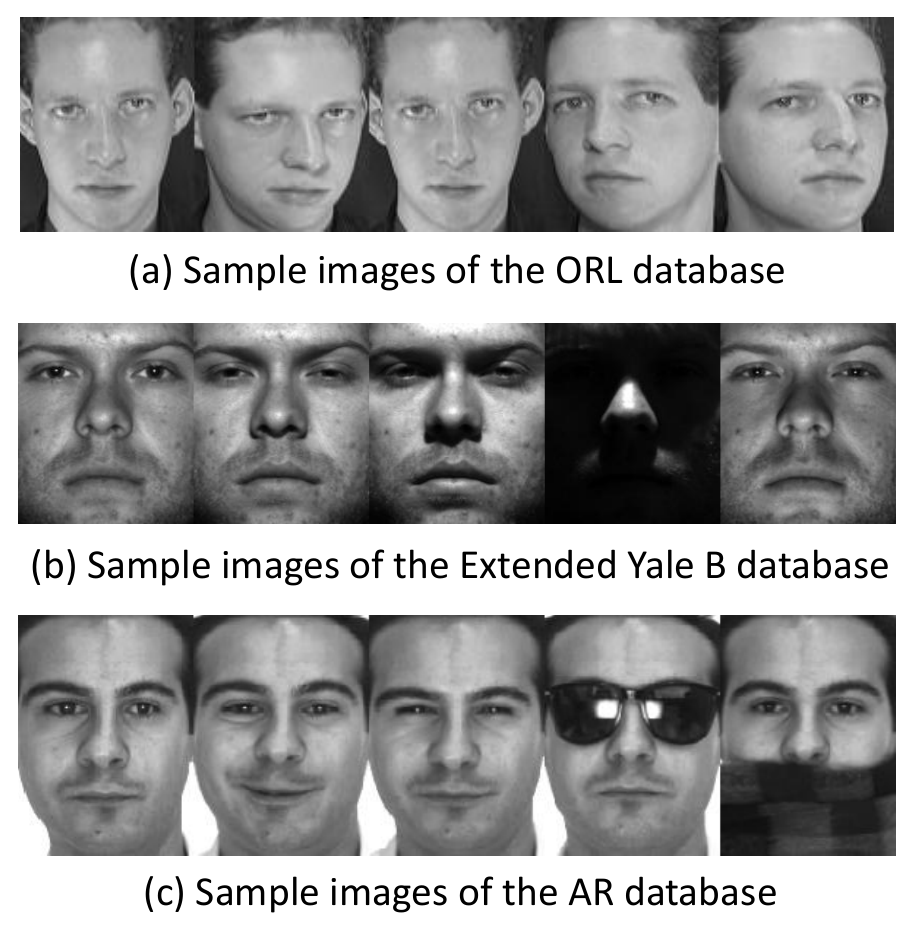}
\caption{Example images of different face database used in our experiments.}
\label{fig:exam_img}
\end{figure}

The ORL database \cite{samaria1994parameterisation} contains 400 face images of 40 people in total, 10 images for each subject. The face images are taken under different lighting conditions, varying facial expressions and facial details, some example images are shown in Fig.~\ref{fig:exam_img} (a). Following the protocol in \cite{li2014learningb}, for each class, we randomly split the data into two parts, one part as the training images, and the other part as the testing images. The size of the images are cropped to 28$\times$23. All the face images are manually corrupted by an unrelated block image at a random location. The learned dictionary contains 5 atoms for each class.  Generally, the parameters are set as follows, $\lambda=0.05$, $\beta=0.1$ and $\gamma=0.0001$ on the ORL database with different levels of block corruptions. We repeat the experiments 10 times and the average recognition accuracy is listed in Table \ref{table:orl_res}.

\begin{table}[h]
\centering
\caption{Recognition accuracy (\%) on the ORL database with different levels of occlusion (\%).}
\label{table:orl_res}
\begin{tabular}{ccccccc}
\hline
Occlusions & 0 & 10 & 20 & 30 & 40 & 50\\ \hline
LRRID  &  94.85 & 92.20 & 90.60 & 86.80 & 82.10	& 77.65   \\
LRRS~\cite{li2014learningb}  &     96.05 & 93.25 & 90.85 & 86.60 & 77.50	& 70.90  \\
RCBD~\cite{li2014learningb}   &  95.35 & 93.25 & 92.30 & 89.00 & 81.05	& 73.10  \\
DLRD\_SR~\cite{ma2012sparse} &   95.60 & 92.60 & 90.75 & 85.50 & 79.20	& 70.95  \\
SRC~\cite{wright2008robust}     &  96.45 & 91.25 & 83.20 & 73.00 & 57.25 & 50.15 \\ \hline
\end{tabular}
\end{table}

As we can see from Table \ref{table:orl_res}, compared to other methods in the same condition, although our method (LRRID) does not have the highest recognition accuracy all the time, our method shows high robustness to the severe corruption like block occlusion, and achieves significant improvement. We can find that our method has stronger identification capability than LRRS by utilizing the incoherent dictionary. The recognition accuracy shows that the performance of our method is better when both training and testing face images are badly corrupted. To some extent, it validates that the incoherent dictionary contains more discriminating information. It also demonstrates that high quality dictionary is demanded for learning discriminative representation when both training and testing face images are badly corrupted.

\begin{figure}[h]
\centering
\includegraphics[trim={0mm 0mm 0mm 0mm},clip, width = .8\textwidth]{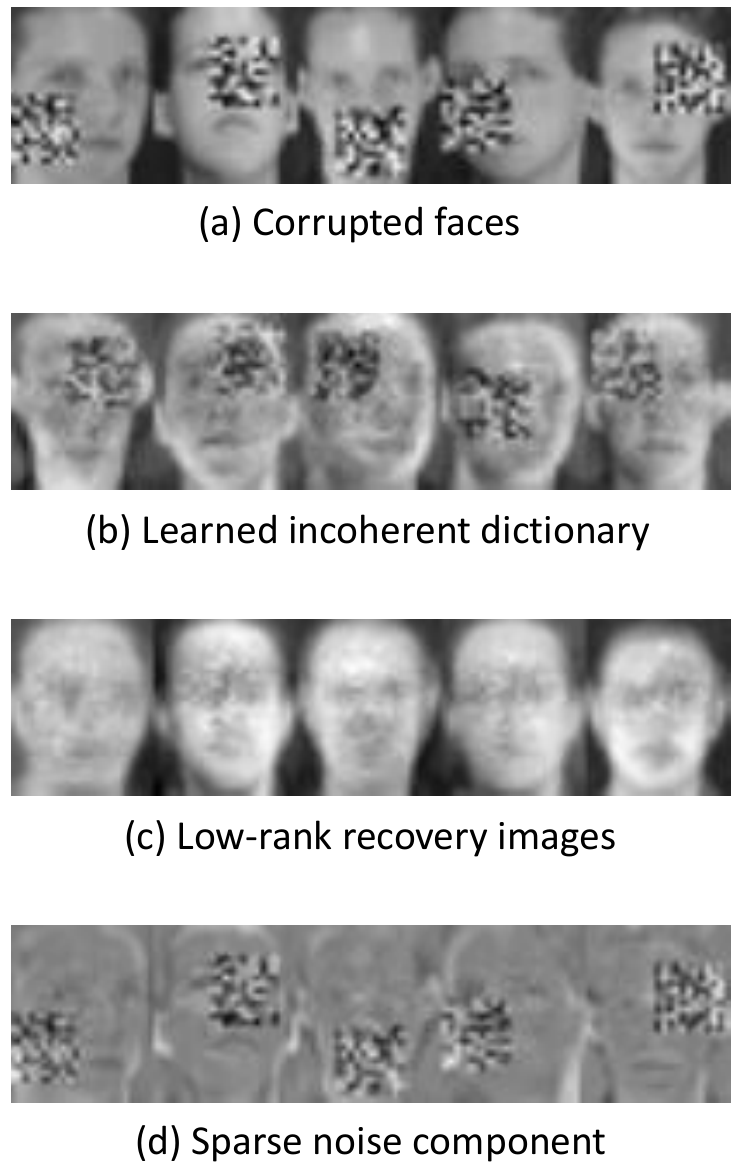}
\caption{Decomposition results of our method on the ORL database with 20\% random block occlusion.}
\label{fig:orl_recover}
\end{figure}

We present some examples of decomposition results in Fig.~\ref{fig:orl_recover} with 20\% block occlusion. As shown in Fig.~\ref{fig:orl_recover}, the top five images are original images with 20\% block corruption, while the second row is the learned incoherent dictionary ($\mathbf{D}$) corresponding to the testing images, the third row denotes the low-rank recovery images ($\mathbf{DX}$) and the bottom five images are the corresponding sparse noise ($\mathbf{E}$) including block corruption, facial expression, and pose changes.

\subsection{Extended Yale B }
The Extended Yale B \cite{georghiades2001few,lee2005acquiring} dataset contains 38 subjects with 2414 face images, they are taken under different lighting conditions, Fig.~\ref{fig:exam_img} (b) presents some example images. For each class, there are between 59 and 64 images. As in Ref. \cite{ma2012sparse}, we randomly select 32 images from each subject as the training samples, and the rest images as the testing samples. All the methods are conducted on the down-sampled images with various ratios of 1/32, 1/24 and 1/16. The learned dictionary contains 32 atoms for each class. And the parameters are set as follows, $\lambda=0.1$, $\beta=0.1$ and $\gamma=0.0001$ on the Extended Yale B database. We perform 10 times for each experiment and the average recognition accuracy is shown in Table 2. 

\begin{table}[h]
\centering
\caption{Recognition accuracy (\%) on Extended Yale B database.}
\label{table:eyaleb_res}
\begin{tabular}{cccc}
\hline
Dimensions & 30 & 56 & 120 \\ \hline
LRRID           & 78.26 & 90.44 & 96.41       \\
LRRS~\cite{li2014learningb}         & 77.21 & 87.03 & 95.38       \\
RCBD~\cite{li2014learningb}         & 53.56 & 69.27 & 82.61       \\
DLRD\_SR~\cite{ma2012sparse}       & 67.60 & 84.76 & 93.24   \\
SRC~\cite{wright2008robust}        & 79.10 & 88.03 & 93.15      \\ \hline
\end{tabular}
\end{table}

We can see that our approach achieves the best results and outperforms the state-of-the-art method RCBD significantly. The method of RCBD does not have a superb performance when the dimensionality is low, it may lose some structure information. The recognition accuracy of DLRD\_SR is not very high when the dimension is low, while our method has better performance. After dictionary learning, our method outperforms LRRS by 1.83\% on average. The experimental results validate the effectiveness and robustness of our method when face images are corrupted by illumination variations.

\subsection{AR Face Database}
The AR face database \cite{martinez2001pca} includes over 4000 face images for 126 individuals. There are 26 images for each individual, and they are taken in two separated sessions. There are 13 images of each person for each season, in which 3 images are occluded by sunglasses, another 3 with scarves, and the remaining 7 with different facial expressions or illumination variations (and thus they are named as unobscured images), each image is 165$\times$120 pixels, some example images are depicted in Fig.~\ref{fig:exam_img} (c). As in \cite{li2014learningb}, we perform the experiments on the down-sampled images with the ratio of 3$\times$3, so the dimensionality of sample features is 2200. Experiments are conducted under the following three different scenarios.

\textit{Sunglasses}: In this scenario, we take into account the situation that both training and testing samples are corrupted by sunglasses occlusion. 7 unobscured images and 1 image with sunglasses (which is randomly chosen) from the first session are taken as the training samples, and 7 unobscured images from the other session and the remaining 5 images with sunglasses from two sessions are taken as the testing samples. Sunglasses generate about 20\% occlusion of the face image.

\textit{Scarf}: In this scenario, we take into account the situation that both training and testing samples are corrupted with the scarf occlusion. Similarly, 8 training images consist of 7 unobscured images plus 1 image with scarf (which is randomly chosen) from the first session, and 12 testing images consist of 7 unobscured images from the other session and the remaining 5 images with scarf from two sessions. Scarf generates about 40\% occlusion of the face image.

\textit{Mixed (Sunglasses+Scarf )}: In this scenario, we consider the case when both training and testing samples are occluded by sunglasses or scarf. 7 unobscured images and 2 occluded images (one is randomly chosen from the images with sunglasses and the other one is randomly chosen from the images with scarf) from the first session are taken as the training samples, and the remaining 17 images in two sessions are used as testing samples.

Following the protocol in \cite{li2014learningb}, a compact dictionary with 5 items for each class is preferred under different scenarios. And the parameters are set as follows, $\lambda=0.1$, $\beta=0.1$ and $\gamma=0.0001$ on the AR database. We perform 10 times for each experiment and the average recognition accuracy is listed in Table~\ref{table:occlusion}. 
\begin{table}[h]
\centering
\caption{Recognition accuracy (\%) on the AR database.}
\label{table:occlusion}
\begin{tabular}{cccc}
\hline
Scenario & Sunglasses & Scarf & Mixed \\ \hline
LRRID           & 94.35 & 91.74 & 91.82      \\
LRRS~\cite{li2014learningb}      & 90.51 & 89.29 & 87.57     \\
RCBD~\cite{li2014learningb}       & 94.02 & 92.13 & 91.79      \\
DLRD\_SR~\cite{ma2012sparse}      & 67.02 & 72.69 & 62.69  \\
SRC~\cite{wright2008robust}        & 92.45 & 91.63 & 90.87    \\ \hline
\end{tabular}
\end{table}

Our method has the highest recognition accuracy under different scenarios. High quality dictionary is demanded for learning discriminative representations, LRRS and SRC* cannot handle the situations well, both of them choose part of the training samples as the dictionary. Different from them, our method and RCBD learn a compact and discriminative dictionary, which can improve the performance. Compared with LRRS, the results verify the incoherent dictionary is a high quality dictionary. The results from Table~\ref{table:occlusion} demonstrate that our method is robust to real disguise in the face images.

\begin{figure}[h]
\centering
\includegraphics[trim={0mm 0mm 0mm 0mm},clip, width = .8\textwidth]{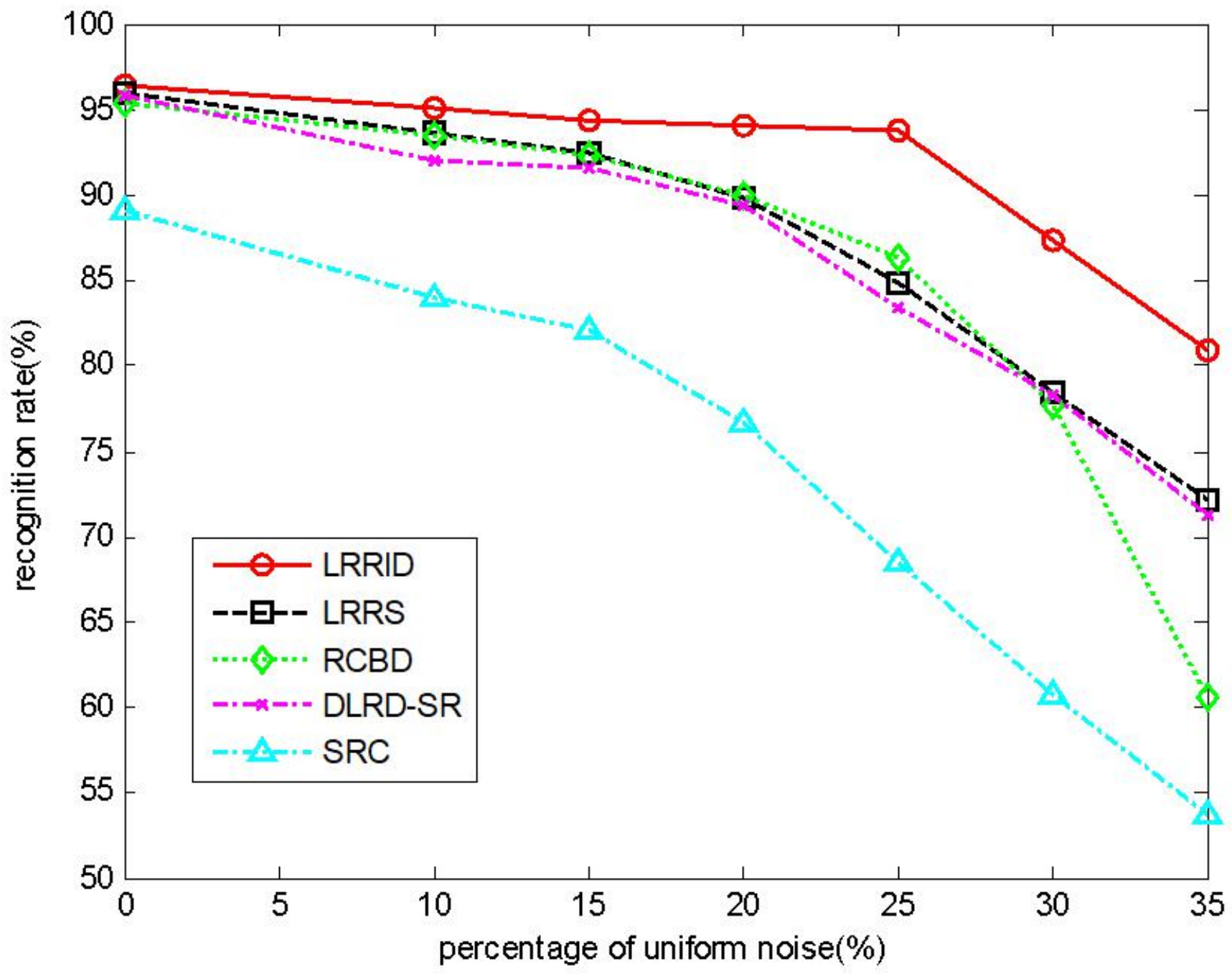}
\caption{Recognition accuracy on the AR database with different levels of pixel corruption.}
\label{fig:ipixel_corr}
\end{figure}

To further evaluate the performance of our method on the corrupted data with uniform noise, as protocol in \cite{li2014learningb}, we design a test on the AR database with uniform noise. 7 unobscured images from the first session are taken as the training samples, and 7 another unobscured images from the other session are taken as the testing samples. A percentage of randomly chosen pixels from all the training and testing images are replaced with samples from a uniform distribution over $[0,V_{max}]$ as \cite{li2014learningb} did, where $V_{max}$ is the largest possible pixel value in the image. The learned dictionary contains 7 atoms for each class. Fig.~\ref{fig:ipixel_corr} plots the recognition accuracy against different levels of noise.

From Fig.~\ref{fig:ipixel_corr}, it is clear that the recognition accuracy of all methods decreases with the increasing of noise. Our method performs well and better than RCBD and LRRS. The results show our method is robust to severe uniform noise.

\begin{figure}[h]
\centering
\includegraphics[trim={0mm 0mm 0mm 0mm},clip, width = .8\textwidth]{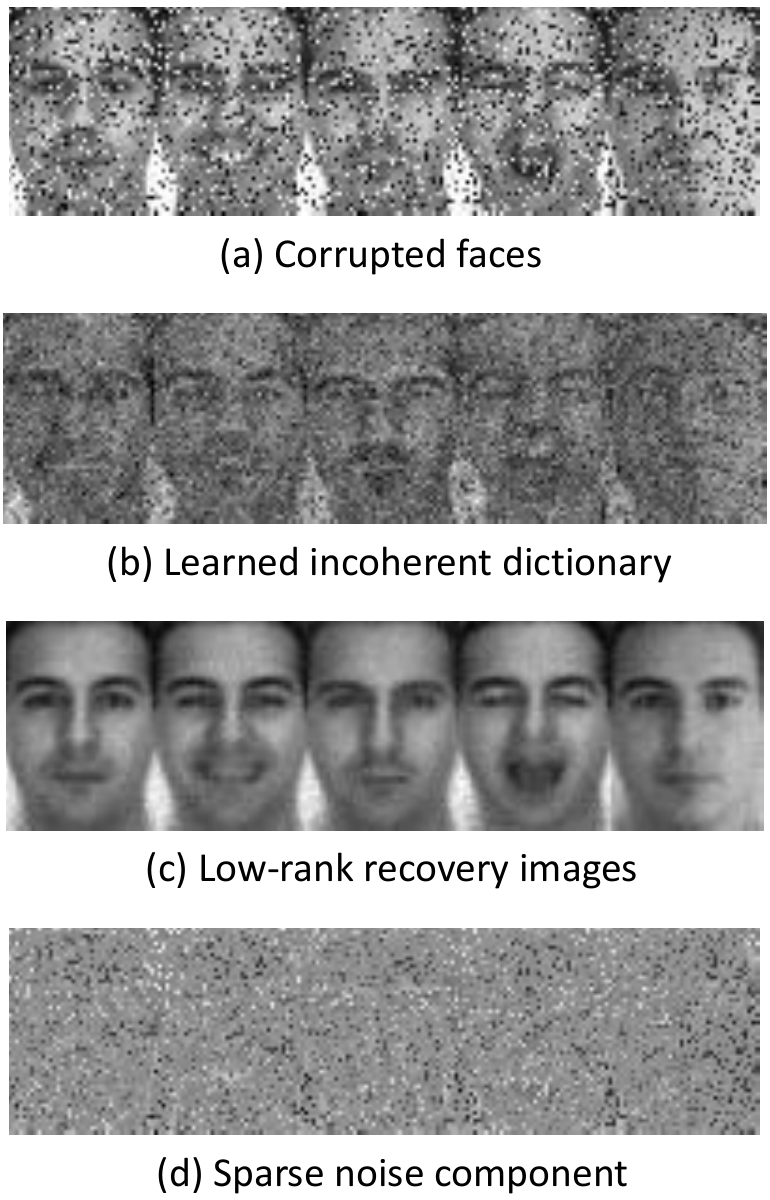}
\caption{Decomposition results of testing images from the AR database with 30\% uniform noises.}
\label{fig:ar_corr}
\end{figure}

Some image decomposition examples of testing images from the AR database are presented in Fig.~\ref{fig:ar_corr}. The first row is the original face images with 30\% uniform noise. The second row is the learned incoherent dictionary ($\mathbf{D}$) corresponding to the testing images, and the third row is the low-rank recovery images ($\mathbf{DX}$), and the bottom row is the noise component ($\mathbf{E}$).

\begin{figure}[h]
\centering
\includegraphics[trim={0mm 0mm 0mm 0mm},clip, width = .8\textwidth]{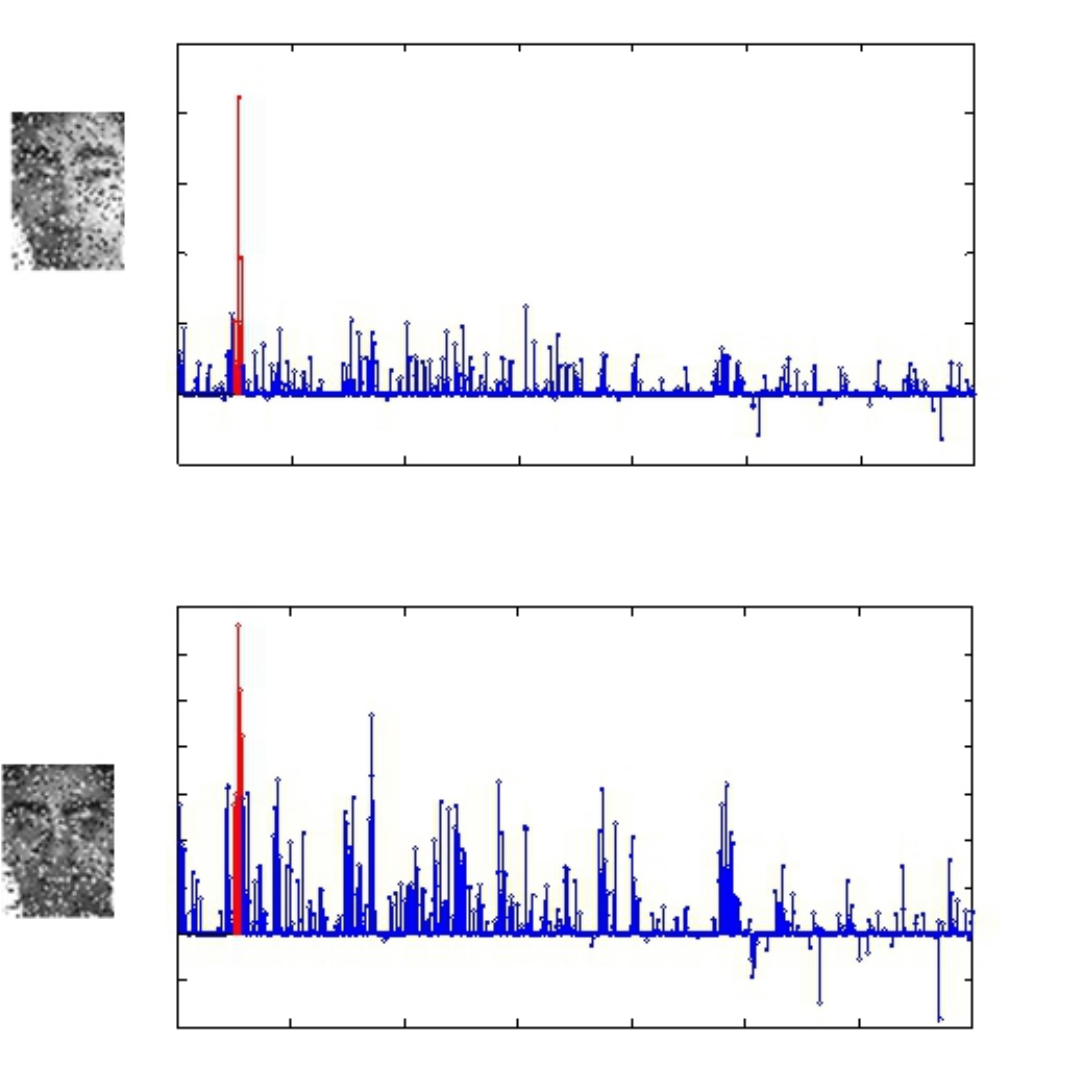}
\caption{Our proposed method represents a training face image and a test face image, the face images are from the AR face database which contain 30\% uniform noise. Red coefficients correspond to the atoms of the correct individual from the learned dictionary $\mathbf{D}$.}
\label{fig:coeff}
\end{figure}

Fig.~\ref{fig:coeff} shows a successful example by using face recognition as an example, the upper part is the example of a training sample and the bottom part is the example of a testing sample. The red coefficients correspond to the atoms of the correct individual from the learned dictionary $\mathbf{D}$, and the blue coefficients belong to the other classes. Fig.~\ref{fig:coeff} demonstrates that the face images can be properly represented, and it also validates the effectiveness of the learned dictionary by our method. 

\section{Conclusion}
\label{sec_8}
In this paper, we proposed a new face recognition method based on the semi-supervised framework, in which the incoherent dictionary is introduced to simultaneously learn the discriminative low-rank and sparse representations for both training and testing images. The research indicates that the proposed method can learn robust representations of training and testing images simultaneously even when both training and testing images are badly corrupted. Furthermore, the new method integrates the correlation penalty into the dictionary learning model which can make the dictionary more discerning. All the updating formulas are derived by using the inexact ALM algorithm. The experimental results confirm that our proposed method is effective and robust, achieving state-of-the-art performance especially when both the training and testing samples are contaminated, including illumination changes, occlusion and pixel corruption. 

Parameter selection is a nontrivial task in pattern classification, and there are some parameters in our proposed method, how to choose the best setting parameters in our method is a problem to be further investigated in our future work. 

\section*{Acknowledgements}
This work was supported by the National Natural Science Foundation of China (Grant No. 61672265, U1836218), the 111 Project of Ministry of Education of China (Grant No. B12018), the Postgraduate Research \& Practice Innovation Program of Jiangsu Province under Grant No. KYLX\_1123, the Overseas Studies Program for Postgraduates of Jiangnan University and the China Scholarship Council (CSC, No.201706790096).

\section*{References}

\bibliography{mybibfile}

\end{document}